\documentclass{article}
\usepackage{ijcai17}

\usepackage{times}

\usepackage{latexsym} 

\usepackage{url}
\usepackage[dvipdfmx]{graphicx}
\usepackage{amssymb}
\usepackage{amsmath}
\usepackage{multirow}
\usepackage{makecell}
\usepackage{colortbl}
\usepackage{CJK}
\usepackage[figuresright]{rotating}
\usepackage{soul}
\usepackage{algorithm}
\usepackage[noend]{algpseudocode}

\title{Supervised Syntax-based Alignment between English Sentences \\ and Abstract Meaning Representation Graphs}

\author{Chenhui Chu$^1$ \and Sadao Kurohashi$^2$\\
  $^1$Japan Science and Technology Agency \\
  $^2$Graduate School of Informatics, Kyoto University \\
  {\tt chu@pa.jst.jp, kuro@i.kyoto-u.ac.jp}}

\begin{document}

\maketitle

\begin{abstract}
As alignment links are not given between English sentences and Abstract Meaning
Representation (AMR) graphs in the AMR annotation, automatic alignment becomes
indispensable for training an AMR parser. Previous studies formalize it as a 
string-to-string problem and solve it in an unsupervised way, which suffers from 
data sparseness due to the small size of training data for English-AMR alignment. 
In this paper, we formalize it as a syntax-based alignment problem and solve it 
in a supervised manner based on syntax trees, which can address the data
sparseness problem by generalizing English-AMR tokens to syntax tags.
Experiments verify the effectiveness of the proposed method not only for 
English-AMR alignment, but also for AMR parsing.
\end{abstract}

\section{Introduction}
\begin{figure}
\centerline{\includegraphics[width=\hsize]{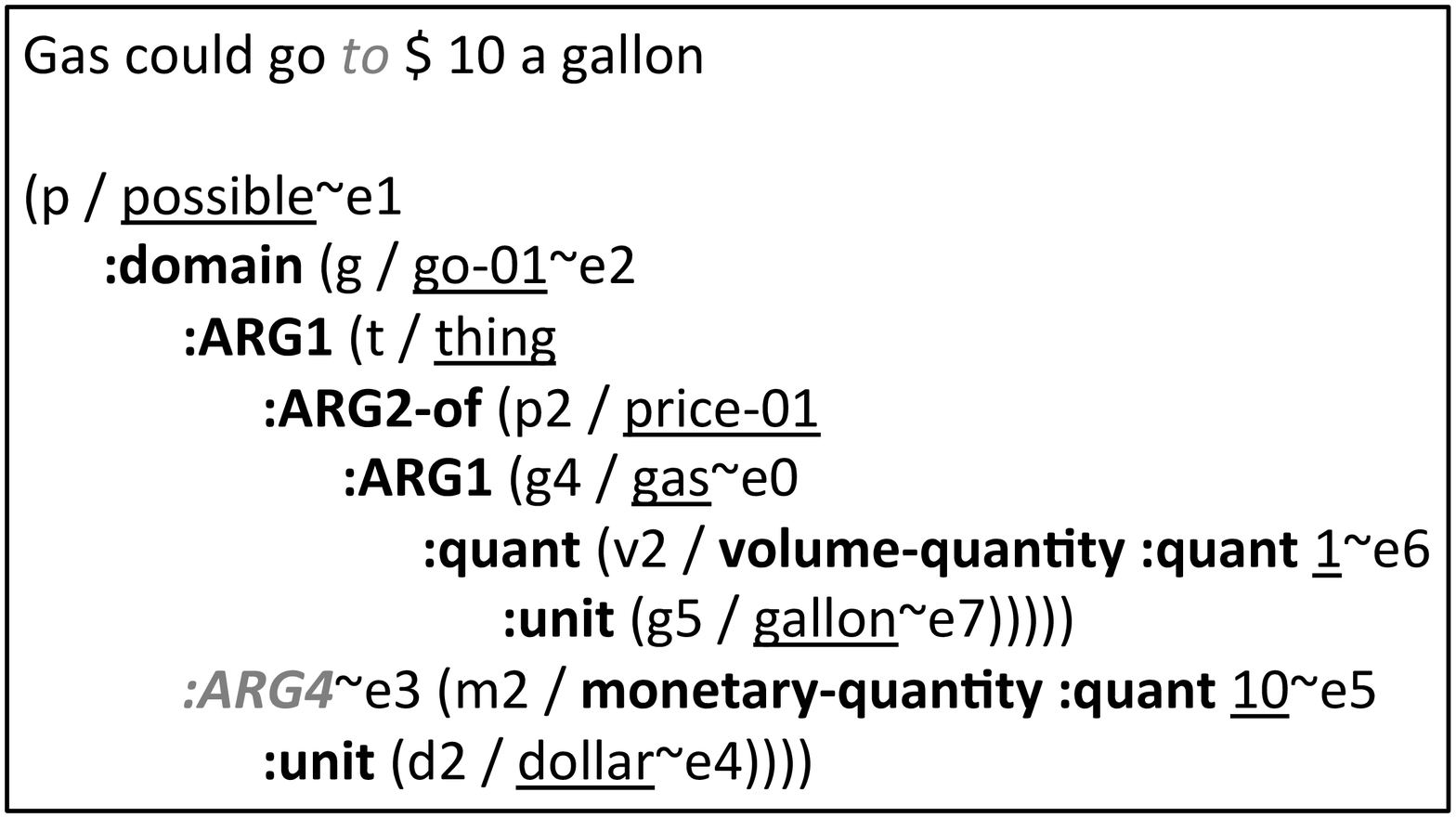}}
\caption{An example of English-AMR alignment (AMR concepts are underlined, 
AMR roles are in bold, $\sim$e denotes alignment and the numbers after
$\sim$e are English token indices).}
\label{fig:amr_alignment}
\end{figure}

Abstract Meaning Representation (AMR) is a sentence level semantic annotation, 
which is represented in a rooted, directed, and edge-labeled graph 
\cite{banarescu-EtAl:2013:LAW7-ID}. Nodes of a graph are {\it concepts} 
(e.g., ``possible'' in Figure \ref{fig:amr_alignment}), while edges are 
labeled with semantic {\it roles} (e.g., ``:ARG4'' in Figure \ref{fig:amr_alignment}). 
AMR concepts consist of predicate senses, named entities, and lemmas of English tokens. 
AMR roles consist of core semantic roles from the Propbank \cite{Palmer:2005:PBA:1122624.1122628}
and fine-grained semantic roles defined specifically for AMR.
As AMR annotation has no explicit alignment with the tokens in the English sentence,
automatic alignment becomes a requirement for training AMR parsers
\cite{flanigan-EtAl:2014:P14-1,wang-xue-pradhan:2015:ACL-IJCNLP,werling-angeli-manning:2015:ACL-IJCNLP,artzi-lee-zettlemoyer:2015:EMNLP,pust-EtAl:2015:EMNLP,zhou-EtAl:2016:EMNLP20163,misra-artzi:2016:EMNLP2016,Peng:2017:EACL}.

The alignment problem between English sentences and AMR graphs is not
trivial. There are two reasons. Firstly, the problem itself is complicated.
Concepts do not always have a direct matching among the English tokens in a given sentence.
For example, in Figure \ref{fig:amr_alignment} the English token ``could'' 
is represented as the concept ``possible,'' and aligning them is not easy. 
It becomes more difficult in the English token-to-role alignment case. For example,
in Figure \ref{fig:amr_alignment}, we should align the English token ``to'' to the role ``:ARG4.''
Secondly, the training data for English-AMR alignment is very small. The biggest
publicly available data set only contains 13,050 English-AMR pairs, which is significantly
smaller than conventional alignment settings in machine translation (MT).

\cite{flanigan-EtAl:2014:P14-1} is the first study of English-AMR alignment, 
which proposes a rule-based method. The limitation of the rule-based method is that it cannot
benefit from more annotation of English-AMR pairs. A data driven alignment
method also has been proposed \cite{pourdamghani-EtAl:2014:EMNLP2014}.
They formalize the string-to-graph alignment problem as a string-to-string 
problem by linearizing the AMR graph. Then they apply the conventional 
unsupervised string-to-string alignment models (i.e., IBM models \cite{brown-EtAl:1993}) 
for this problem. However, this method significantly suffers from data
sparseness due to the small size of training data.

Using syntax trees in a supervised manner has shown its effectiveness
in the alignment problem in MT \cite{riesa-irvine-marcu:2011:EMNLP}. 
Motivated by this, in this paper, we formalize the English-AMR graph 
alignment problem as a syntax-based alignment problem. We then apply the 
supervised syntax-based alignment model \cite{riesa-irvine-marcu:2011:EMNLP}
for this. Our proposed method generalizes pure English-AMR tokens to syntax
tags, which can address the data sparseness problem of the previous
study \cite{pourdamghani-EtAl:2014:EMNLP2014}.

Experiments conducted on the benchmark dataset show that our proposed method outperforms 
the unsupervised alignment model of \cite{pourdamghani-EtAl:2014:EMNLP2014} 
by 1.7\% absolute F-score on alignment accuracy, although it is trained on only 100 
English-AMR pairs that are annotated with gold alignments. Using the alignments
by our proposed method instead of the unsupervised alignments for a state-of-the-art 
AMR parser \cite{pust-EtAl:2015:EMNLP} improves the parsing accuracy 
of 0.4\% absolute Smatch F-score \cite{cai-knight:2013:Short}.

\section{Related Work}
Three different alignment criteria, and their corresponding alignment methods have been proposed
\cite{flanigan-EtAl:2014:P14-1,pourdamghani-EtAl:2014:EMNLP2014,werling-angeli-manning:2015:ACL-IJCNLP}.
\cite{flanigan-EtAl:2014:P14-1} is the most prior work of English-AMR alignment. They proposed 
an alignment criterion that aligns a span of English tokens to an AMR graph concept fragment. This means 
that in Figure \ref{fig:amr_alignment}, e.g., the English token ``gas'' should be aligned 
to the graph fragment ``(t / thing :ARG2-of (p2 / price-01 :AMR1 (g4 / gas.'' 
Their criterion, however, does not align AMR roles explicitly. \cite{flanigan-EtAl:2014:P14-1} 
proposed a rule-based method for this type of alignment. 
In contrast, \cite{pourdamghani-EtAl:2014:EMNLP2014} proposed an alignment criterion that
not only aligns AMR concepts but also roles to English tokens explicitly. In addition, in their criterion,
each concept and role is aligned to at most one English token, but each English token
can be aligned to many concepts/roles. An example of the alignment
criterion of \cite{pourdamghani-EtAl:2014:EMNLP2014} is shown in Figure \ref{fig:amr_alignment}.
They proposed an unsupervised string-to-string alignment model for this.
\cite{werling-angeli-manning:2015:ACL-IJCNLP} essentially adopted the alignment criterion
of \cite{pourdamghani-EtAl:2014:EMNLP2014}, except that they forced every AMR concept aligns
to some English tokens. They proposed a boolean linear programming method for this. 
Note that the accuracies of these studies were reported on different golden alignment data, 
and thus they are not directly comparable.
We adopt the alignment criterion of \cite{pourdamghani-EtAl:2014:EMNLP2014}, and
directly compare the alignment accuracy with their method.

For AMR parsing, \cite{flanigan-EtAl:2014:P14-1} is also the most prior work. They proposed a graph 
based parsing method that finds a maximum spanning and connected subgraph via structured prediction 
for AMR parsing. Their parser is publicly available as the JAMR parser.\footnote{http://github.com/jflanigan/jamr}
\cite{werling-angeli-manning:2015:ACL-IJCNLP} extended the study of \cite{flanigan-EtAl:2014:P14-1} by 
proposing generative actions for subgraph derivation based on their alignment criterion.
\cite{zhou-EtAl:2016:EMNLP20163} extended the study of \cite{flanigan-EtAl:2014:P14-1} 
by proposing a beam search algorithm.
\cite{wang-xue-pradhan:2015:ACL-IJCNLP} proposed a transition based method that first
parses English sentences to dependency trees and then transforms the dependency trees 
to AMR graphs. Their parser is publicly available as the CAMR parser.\footnote{https://github.com/c-amr/camr}
\cite{artzi-lee-zettlemoyer:2015:EMNLP} proposed using the combinatory categorial grammar (CCG) for AMR parsing.
\cite{misra-artzi:2016:EMNLP2016} developed the CCG AMR parsing method of \cite{artzi-lee-zettlemoyer:2015:EMNLP} based on neural networks.
Note that \cite{flanigan-EtAl:2014:P14-1,wang-xue-pradhan:2015:ACL-IJCNLP,artzi-lee-zettlemoyer:2015:EMNLP,zhou-EtAl:2016:EMNLP20163,misra-artzi:2016:EMNLP2016}
adopted the alignment criterion and method of \cite{flanigan-EtAl:2014:P14-1}.
\cite{pust-EtAl:2015:EMNLP} treated AMR parsing as a string-to-tree, syntax-based MT
problem. After transforming English sentences to trees, they further convert the trees to AMR graphs. 
Their parser is publicly available as the ISI AMR parser.\footnote{http://www.isi.edu/$\sim$pust/amrparser.tar.gz}
Seqence-to-seqence based AMR parsing also has been proposed \cite{Peng:2017:EACL}, however it
suffers from data sparseness due to the small size of AMR training data.
Both \cite{pust-EtAl:2015:EMNLP} and \cite{Peng:2017:EACL} adopted the alignment criterion and method of \cite{pourdamghani-EtAl:2014:EMNLP2014}.
Note that the parsers of \cite{werling-angeli-manning:2015:ACL-IJCNLP,zhou-EtAl:2016:EMNLP20163,Peng:2017:EACL} are not publicly available.
We apply the alignments by our proposed method to the ISI AMR parser \cite{pust-EtAl:2015:EMNLP}, 
and also compare the parsing performance with the other publicly available parsers (i.e., JAMR and CAMR).


\section{Baseline Alignment Method}
The baseline method that we compare to is the ISI alignment \cite{pourdamghani-EtAl:2014:EMNLP2014}.
The ISI alignment method formalizes the English-AMR graph alignment problem as
a string-to-string alignment problem, by linearizing an AMR graph to a string. 
It includes three steps: preprocessing, string-to-string alignment, and postprocessing.
\subsection{Preprocessing}
\begin{itemize}
\itemsep=-1mm
\item Linearize the AMR using a depth-first traversal. For example, the AMR graph
in Figure \ref{fig:amr_alignment} will be linearized to ``possible :domain go-01 
:ARG1 thing :ARG2-of price-01 :ARG1 gas :quant volume-quantity :quant 1 :unit 
gallon :ARG4 monetary-quantity :quant 10 :unit dollar.''
\item Remove the tokens that are rarely aligned, to improve the precision with a
small sacrifice of recall. On the English side, this removes stop words, such as articles
``a'', ``an'', ``the''; On the AMR side, this removes special concepts, and roles, such as ``:arg0'', 
``:quant'', ``:op1'' that do not usually align, quotes, and sense tags. 
After this step, the English sentence in Figure \ref{fig:amr_alignment} becomes 
``Gas could go to \$ 10 gallon''; the AMR is transferred to ``possible :domain 
go thing :arg2-of price gas 1 gallon :arg4 10 dollar''.
\item Lowercase and stem both sides to the first four letters. 
This is necessary to address the sparseness of the training data, which is very small
compared to the size of the training data for conventional word alignment of MT. 
This converts English to ``gas coul go to \$ 10 gall'', and AMR to 
``poss :domain go thin :arg2-of pric gas 1 gall :arg4 10 doll'' in Figure \ref{fig:amr_alignment}.
\end{itemize}

\subsection{String-to-String Alignment}
As the preprocessing step has converted the English-AMR graph alignment problem 
to a string-to-string alignment problem,
the widely used IBM alignment models \cite{brown-EtAl:1993} that are based on token 
sequences can be applied. To further improve the alignment accuracy, 
\cite{pourdamghani-EtAl:2014:EMNLP2014} also proposed a symmetrization constraint 
that encourages agreement of the parameter learning in two directions for the IBM models.

\subsection{Postprocessing}
The string-to-string alignments are finally projected back to the original English
sentence and the AMR graph to obtain English-AMR graph alignments. This can be
done easily by memorizing the corresponding token positions before and after the
preprocessing.

\section{Proposed Alignment Method}
We use the same pipeline as \cite{pourdamghani-EtAl:2014:EMNLP2014},
however, we formalize it as a constituency tree based alignment problem, and apply
the hierarchical alignment model of \cite{riesa-irvine-marcu:2011:EMNLP}.
This method has been proposed for conventional word alignment
of MT, however, it has not been used for English-AMR graph alignment.

\subsection{Constituency Trees for English and AMR}
\label{sec:amr2tree}
\begin{figure}
\centerline{\includegraphics[width=\hsize]{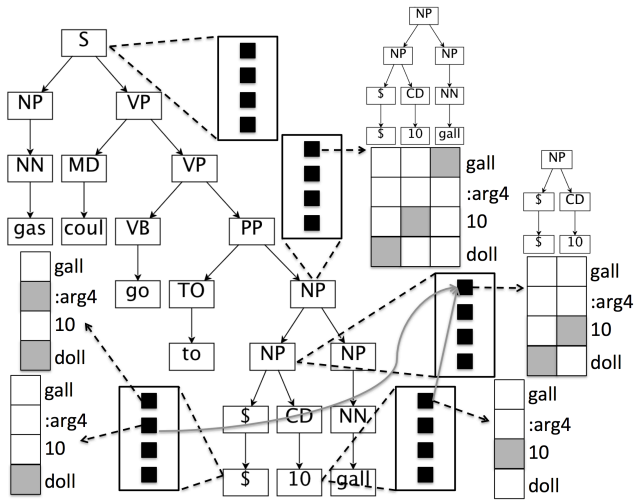}}
\caption{Proposed alignment method (Each black square represents a partial alignment; 
each gray square represents an alignment link in an alignment matrix).}
\label{fig:proposed}
\vspace{-2mm}
\end{figure}

Constituency trees for English can be obtained via a conventional
syntactic parser. In this study, we parse original English sentences with the Berkeley 
parser\footnote{https://github.com/slavpetrov/berkeleyparser} 
\cite{petrov-klein:2007:main}. We process obtained constituency trees by 
discarding the stop words, and replacing the leaf tokens with their stems.
An example of the final tree is shown in Figure \ref{fig:proposed}.

For AMR, we convert AMRs to constituency trees using the method proposed 
in \cite{pust-EtAl:2015:EMNLP} with the following steps: 
\begin{itemize}
\itemsep=-1mm
\item Arbitrarily disconnect multiple parents from each node.
\item Propagate the edge labels (roles) to leaves, and add pre-terminals X.
\item Restructure the tree with role labels as intermediates.
\end{itemize}
We do not apply the reordering steps, because they requires alignments.
For more details of these steps, please refer to \cite{pust-EtAl:2015:EMNLP}.
We used the AMR to syntax tree conversion code provided by \cite{pust-EtAl:2015:EMNLP} 
for the above conversion.

We then process the converted AMR trees by discarding special concepts 
and roles that are rarely aligned, and replacing leaf tokens with their stems.
Note that the converted AMR trees usually are not isomorphic 
to the English trees. For example, ``could'' is the grandchild of the root 
in the English tree, while in the converted AMR tree ``possible'' is
the direct child of the root.


\subsection{Hierarchical Alignment on Constituency Trees}
Figure \ref{fig:proposed} shows an overview of the application of the hierarchical 
alignment model \cite{riesa-irvine-marcu:2011:EMNLP} to our problem.
The model hierarchically searches for the k-best alignment by constructing partial 
alignments over a target constituency tree,\footnote{The source side could be 
either a constituency tree or a token sequence.} in a bottom-up manner (from leaf nodes
to the root). Each node in the tree has partial alignments, which are sorted by 
alignment scores. A partial alignment for a node is an alignment matrix of AMR 
tokens or {\it null}, covered by the node, and it is represented as a black square.
We only keep a beam size of $k$ for partial alignments for each 
node,\footnote{We used a beam size of 128 in our experiments.}
to reduce the computational cost. For example, in Figure \ref{fig:proposed}
the beam size $k=4$. Firstly, 4-best partial alignments are 
generated for all the leaf nodes. These partial alignments are then linearly combined 
to generate partial alignments for the non-terminals in the constituency tree. 
For example, the partial alignments of the leaf node ``\$'' and ``10'' are combined 
to generate 4-best partial alignments for the node ``NP''. We hierarchically 
perform this process until we reach the root node.

One important merit of this model is that it is a discriminative model that can 
incorporate various features including local and non-local features.
Local features are the ones that can be factored among the local
productions in a tree, and otherwise they are non-local features. Local features
include source syntactic, target syntactic, source-target joint syntactic, 
translation rule and same token features. Non-local features include 
lexical translation probabilities, and third party alignment features.

The score of a partial alignment is a linear combination of
these features by their weights. The weights of the features
are learnt against a set of pairs with gold alignments,
using the online averaged perceptron algorithm \cite{collins:2002:EMNLP02}.
The learning objective is defined as:
\begin{equation}
\newcommand{\argmax}{\mathop{\rm arg~max}\limits}
\hat{y}=\argmax_{y \in Y(x_i)}L(y_i,y)+w \cdot h(y)
\end{equation}
where $x_i$ is a token sequence pair and their parse trees; $y_i$ is the 
gold alignment for $x_i$; $Y(x_i)$ denotes all the possible alignment outputs for $x_i$;
$w$ is a weight vector; $h(y)$ is a vector of feature values; $L(y_i,y)$ is a loss
function to measure how bad it would be to guess $y$ instead of $y_i$, 
which is defined as:
\begin{equation}
L(y_i,y)=1-F_1(y_i,y)
\end{equation}
where $F_1(y_i,y)$ is a F-score. $w$ is updated at each iteration as:
\begin{equation}
w=w+h(y_i)-h(\hat{y})
\end{equation}
We stop the update of $w$ when a predefined iteration number has been reached.
After the learning of a set of $w$ at each iteration, we select the $w$ that achieves
the best F-score on a development set for decoding.
For more details of the features and the learning algorithm, please
refer to \cite{riesa-irvine-marcu:2011:EMNLP}. 

\section{Experiments}
We conducted both alignment and AMR parsing experiments, to verify the effectiveness 
of our proposed alignment method.

\begin{table}
\begin{center}
\begin{tabular}{@{}l|r|r|r@{}}\hline
\multicolumn{4}{c}{\bf Original split} \\ 
& \bf train &\bf dev &\bf test \\ \hline \hline
\# pairs & 10,311 & 1,368 & 1,371 \\ 
\# AMR tokens & 364k & 48.9k & 51.0k \\ 
\# AMR roles & 177k & 23.7k & 24.8k \\ 
\# English tokens & 213k & 28.8k & 29.5k \\ \hline
\multicolumn{4}{c}{\bf Our split} \\ 
& \bf train &\bf dev &\bf test \\ \hline \hline
\# pairs & 10,311{\it+200} & 1,368{\it-100} & 1,371{\it-100} \\ 
\# AMR tokens & 370k & 45.1k & 48.6k \\ 
\# AMR roles & 180k & 21.9k & 23.6k \\ 
\# English tokens & 217k & 26.5k & 27.8k \\ \hline
\end{tabular}
\end{center}
\vspace{-3mm}
\caption{\label{table:amr_corpus} Statistics of the AMR corpus for AMR parsing.}
\vspace{+2mm}
\end{table}

\begin{table}
\begin{center}
\begin{tabular}{l|r|r|r}
\hline &\bf train &\bf dev &\bf test \\ \hline \hline
\# pairs & 100 & 50 & 50 \\
\# AMR tokens & 3.0k & 1.5k & 1.6k \\
& (54.7\%) & (51.6\%) & (52.2\%) \\
\# AMR roles & 1.5k & 0.7k & 0.8k \\
& (23.5\%) & (21.0\%) & (21.6\%) \\
\# English tokens & 2.0k & 0.9k & 1.1k \\
& (74.9\%) & (75.8\%) & (75.4\%) \\
\hline
\end{tabular}
\end{center}
\vspace{-3mm}
\caption{\label{table:data} Statistics of the gold alignment data (The numbers in
parentheses are the percentages of the tokens aligned in the gold alignment
data).}
\vspace{-2mm}
\end{table}


\subsection{Settings}
The data used in our experiments was the Linguistic Data Consortium AMR 
corpus release 1.0 (LDC2014T12), consisting of 13,050 AMR/English sentence pairs.\footnote
{https://catalog.ldc.upenn.edu/LDC2014T12} 
The statistics of the original split in the AMR corpus for training, development, and testing
AMR parsers are shown in upper part of Table \ref{table:amr_corpus}.
Among which, 100 development\footnote{http://www.isi.edu/natural-language/mt/dev-gold.txt}  
and 100 testing\footnote{http://www.isi.edu/natural-language/mt/test-gold.txt} pairs were manually 
annotated with gold alignments \cite{pourdamghani-EtAl:2014:EMNLP2014}.
As these 200 pairs were used for training and testing (and tuning for our proposed alignment model) 
alignment models in our study, we moved them to the training data for our AMR parsing experiments.
The statistics of our split of the AMR corpus for AMR parsing is shown in lower part of Table \ref{table:amr_corpus}.
We further mixed these 200 pairs with gold alignments pair by pair, and split them into
100, 50, 50 for training, tuning, and testing, respectively in our alignment experiments.
Table \ref{table:data} shows the statistics of the gold alignment data.



For alignment, we compared our proposed alignment method with the baseline method of \cite{pourdamghani-EtAl:2014:EMNLP2014}.
For the baseline method, we ran the publicly available toolkit ISI 
aligner\footnote{http://www.isi.edu/~damghani/papers/Aligner.zip} on the training data of our split in Table \ref{table:amr_corpus}.
The ISI aligner is an implementation of the method described in \cite{pourdamghani-EtAl:2014:EMNLP2014}.
For our proposed method, we trained and tuned the alignment model on the 100 training and 50 development pairs, 
respectively, with the open source supervised alignment toolkit Nile\footnote{https://github.com/neubig/nile} 
\cite{riesa-irvine-marcu:2011:EMNLP}. As the third party alignment feature for Nile, 
we used the trained ISI alignments. Lexical translation probabilities were generated
from the ISI alignments on the training data of our split in Table \ref{table:amr_corpus}. 
Alignment results were reported on the 50 testing pairs. In addition, we compared different 
ways of using syntax trees for our proposed method:
\begin{itemize}
\itemsep=-1mm
\item AMR(string)-En(tree): Use AMR strings as the source side, and English 
trees as the target side.
\item AMR(tree)-En(tree): Use converted AMR trees as the source side, and English trees 
as the target side.
\item En(tree)-AMR(tree): Use English trees as the source side, and converted AMR trees 
as the target side.
\item Grow-diag-final-and (1): A symmetrization of the alignment results of 
AMR(string)-En(tree) and En(tree)-AMR(tree) with the grow-diag-final-and
heuristic \cite{och--ney:2003}, which is commonly used in MT.
\item Grow-diag-final-and (2): A symmetrization of the alignment results of 
AMR(tree)-En(tree) and En(tree)-AMR(tree) with the grow-diag-final-and
heuristic \cite{och--ney:2003}.
\end{itemize}

\begin{table}
\begin{center}
\begin{tabular}{@{}l|l|r|r|r@{}}
\hline \bf Type & \bf Method &\bf Precision &\bf Recall &\bf F-score \\ \hline \hline
\multirow{3}{*}{Concept} & ISI & 96.2\% & 85.6\% & 90.6\% \\
 & Proposed &\bf 97.1\% &\bf 87.8\% &\bf 92.2\%$\dag$ \\ 
 & Upper bound & 99.7\% & 94.4\% & 97.0\% \\ \hline
\multirow{3}{*}{Role} & ISI & 69.1\% & 43.9\% & 53.7\% \\
 & Proposed &\bf 69.7\% & \bf 47.6\% &\bf 56.6\%$\dag$ \\
 & Upper bound & 95.4\% & 66.1\% & 78.1\% \\ \hline
Concept & ISI & 92.0\% & 77.1\% & 83.9\% \\
+Role & Proposed &\bf 92.7\% &\bf 79.6\% &\bf 85.6\%$\dag$ \\
 & Upper bound & 99.0\% & 88.6\% & 93.5\% \\ \hline
\end{tabular}
\end{center}
\vspace{-3mm}
\caption{\label{table:result} Alignment results (``Proposed'' shows the best 
results among the different syntax usages, ``Upper bound is'' the upper bound 
after removing stop words in English and rarely aligned concepts, roles in AMR,
``$\dag$'' indicates that the F-score is significantly better than ``ISI''
at $p < 0.01$).}
\vspace{-2mm}
\end{table}

\begin{table}
\small
\begin{center}
\begin{tabular}{@{}l@{}|l@{}|r|r|r@{}}
\hline \bf Type & \bf Method &\bf Precision &\bf Recall &\bf F-score \\ \hline \hline
\multirow{4}{*}{Concept} & AMR(string)-En(tree) & 96.1\% & 87.5\% & 91.6\% \\
 & AMR(tree)-En(tree) & 95.9\% & 87.2\% & 91.4\% \\ 
 & En(tree)-AMR(tree) & 94.7\% & 88.0\% & 91.3\% \\ 
 & Grow-diag-final-and (1) & 94.7\% & \bf 88.6\% & 91.6\% \\
 & Grow-diag-final-and (2) &\bf 97.1\% & 87.8\% &\bf 92.2\% \\ \hline
\multirow{4}{*}{Role} & AMR(string)-En(tree) & \bf 78.8\% & 43.3\% & 55.9\% \\
 & AMR(tree)-En(tree) & 77.3\% & 43.3\% & 55.5\% \\
 & En(tree)-AMR(tree) & 61.7\% & 48.6\% & 54.4\% \\ 
 & Grow-diag-final-and (1) & 67.8\% & \bf 50.2\% & \bf 57.7\% \\
 & Grow-diag-final-and (2) & 69.7\% & 47.6\% & 56.6\% \\ \hline
 & AMR(string)-En(tree) &\bf 93.8\% & 78.5\% & 85.5\% \\
Concept & AMR(tree)-En(tree) & 93.4\% & 78.3\% & 85.2\% \\
+Role & En(tree)-AMR(tree) & 88.8\% & 80.0\% & 84.2\% \\ 
 & Grow-diag-final-and (1) & 90.2\% & \bf 80.8\% & 85.2\% \\
 & Grow-diag-final-and (2) & 92.7\% & 79.6\% &\bf 85.6\% \\ \hline
\end{tabular}
\end{center}
\vspace{-3mm}
\caption{\label{table:comparison} Syntax usage comparison results for our proposed method.}
\vspace{-2mm}
\end{table}

For AMR parsing, we compared the parsing performance of the state-of-the-art 
AMR parser \cite{pust-EtAl:2015:EMNLP} using either the baseline ISI alignments
or our proposed alignments, respectively. We trained the baseline
alignment model using the same method described above. Alignments for the parsing 
training data are available once we trained the baseline alignment model. For our proposed 
alignment method, we trained the alignment model with the best syntax usage and further 
applied the trained model on the parsing training data for obtaining alignments.
In addition, we compared with the performance of other public available AMR parsers 
\cite{flanigan-EtAl:2014:P14-1,wang-xue-pradhan:2015:ACL-IJCNLP}
that use alignments with different annotation criteria from ours, by running their parsers with
default settings. All AMR parsing experiments were conducted on our split of the AMR corpus 
in Table \ref{table:amr_corpus}.

\subsection{Alignment Results}
Table \ref{table:result} shows the alignment results. We report the alignment
accuracies for the concept, role and both types of tokens, respectively. 
The significance tests were performed using the bootstrapping method \cite{zhang:755:2004:lrec2004}.
We can see that our proposed method outperforms the ISI alignment for all the alignment types.
However, there is still a gap between it and the upper bound, especially for role alignment.
Although they are not directly comparable, the concept precision 97.1\% and F-score 92.2\% obtained 
by our proposed method are also significantly higher than the concept precision 83.2\%
reported in \cite{werling-angeli-manning:2015:ACL-IJCNLP} and the concept F-score 90\% 
reported in \cite{flanigan-EtAl:2014:P14-1}, respectively.



Table \ref{table:comparison} shows the results of using syntax trees in 
different ways for the proposed method. AMR(tree)-English(tree) performs slightly
worse than AMR(string)-English(tree), due to the bad isomorphism between
converted AMR trees and English trees. Using converted AMR trees as the target 
side seems to be a bad idea for the precision, but it is helpful for improving the recall.
The reason for this is that
English-to-AMR is a one-to-many alignment problem, while En(tree)-AMR(tree)
could produce many-to-one alignments for English-to-AMR due to the peculiarities of Nile, 
which decreases the precision but improves the recall.
The grow-diag-final-and heuristic leverages both the high precision of AMR(tree)-En(tree) 
and the high recall of En(tree)-AMR(tree), and thus further improves the F-score; it also
improves the role F-score by using the grow-diag-final-and heuristic for AMR(string)-En(tree)
and En(tree)-AMR(tree).


\begin{figure}
\centerline{\includegraphics[width=\hsize]{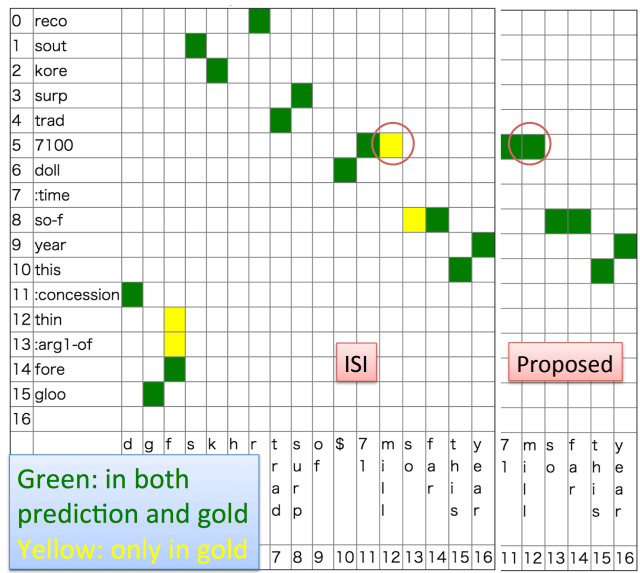}}
\vspace{-3mm}
\caption{An alignment comparison example of better concept recall.}
\vspace{-2mm}
\label{fig:alignment_example_concept_recall}
\end{figure}

To further understand the reason for the alignment improvement, we investigated 
many alignment examples. We found that the main reason for the alignment improvement
is the generalization of pure English-AMR tokens to syntax tags, which addresses
the data sparseness problem. 
Figure \ref{fig:alignment_example_concept_recall} shows 
an alignment example of better concept recall by our proposed method. 
Comparing the ISI alignment and our proposed method, we
can see that our proposed method successfully aligns the AMR concept ``7100(000)''
to the English tokens ``71'' and ``mill(ion),'' while the ISI alignment fails.
``7100(000)'' is not aligned to ``mill(ion),'' due to the sparseness of the training
data. With the help of syntax information, both ``7100(000)'' and ``mill(ion)''
are generalized to cardinal numbers ``Aquant'' and ``QP (CD),'' respectively. As cardinal 
numbers co-occur many times in the training data, our proposed method learns a big 
positive weight, which successfully aligns them.

\begin{figure}
\centerline{\includegraphics[width=\hsize]{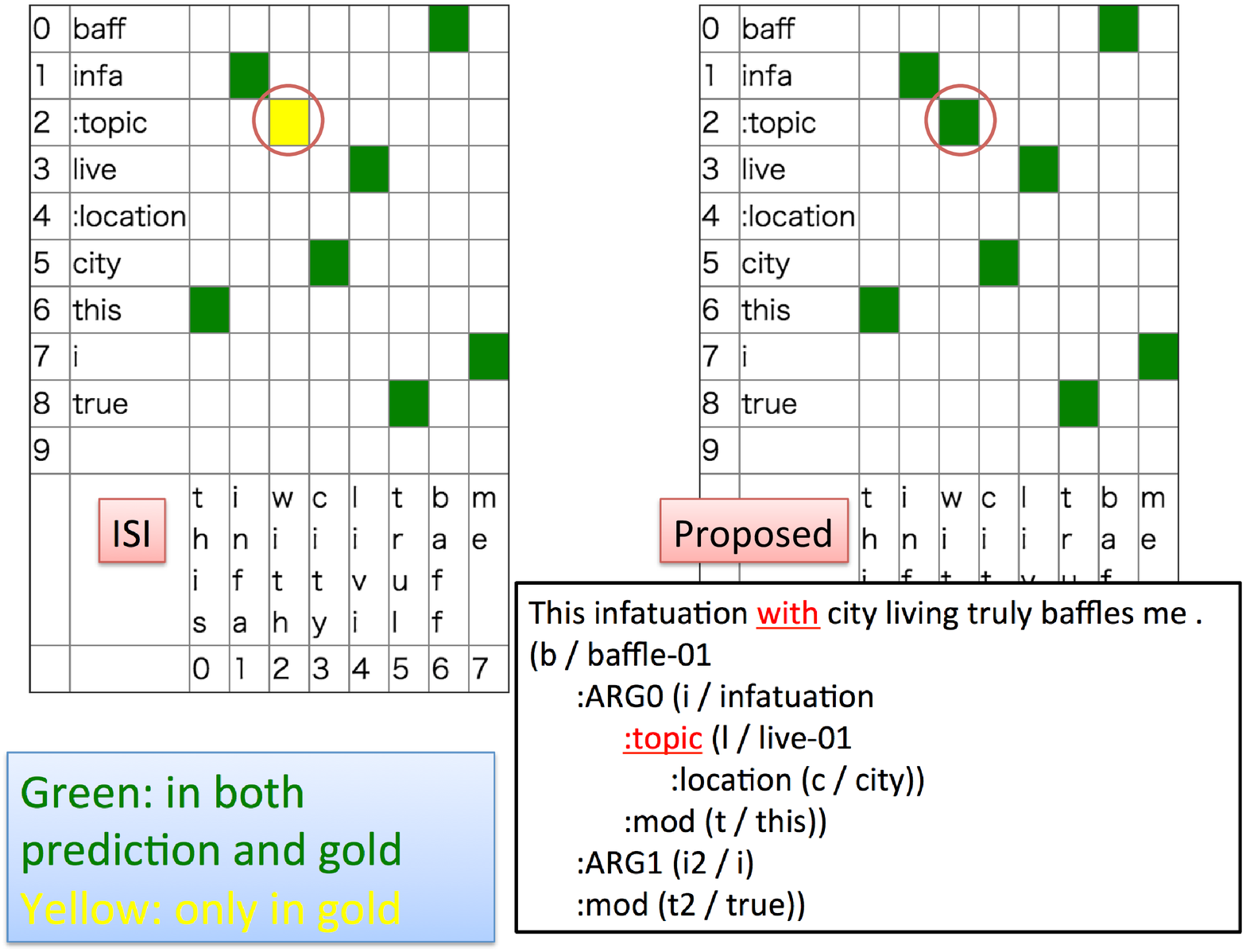}}
\vspace{-3mm}
\caption{An alignment comparison example of better role recall.}
\label{fig:alignment_example_role_recall}
\end{figure}

Figure \ref{fig:alignment_example_role_recall} shows 
an alignment example of better role recall by our proposed method. 
We can see that our proposed method successfully aligns the AMR  role
token ``:topic'' to the English token ``with,'' while the ISI alignment
fails. This is again improved by the syntax information. As ``:topic''
and ``with'' do not co-occur in the training data, it is very difficult for
the ISI alignment model to align them. Our proposed method generalizes
them to ``Ctopic'' and ``IN,'' respectively. For which, a positive alignment 
weight has been learnt, which makes them being aligned.

\begin{figure}
\centerline{\includegraphics[width=\hsize]{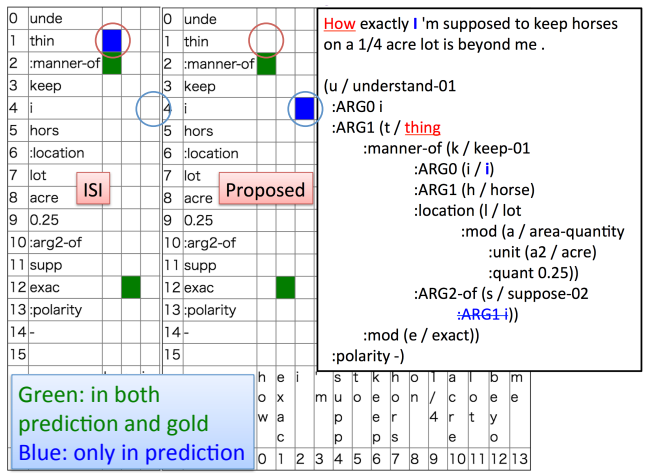}}
\vspace{-3mm}
\caption{An alignment comparison example of concept precision.}
\vspace{-2mm}
\label{fig:alignment_example_role_precision}
\end{figure}

Figure \ref{fig:alignment_example_role_precision} shows an alignment example 
of concept precision comparing the baseline with our proposed method. 
Although the ISI alignment incorrectly aligns the AMR concept token 
``thin(g)'' to the English token ``how,'' our proposed method does not align them.
Because ``thin(g)'' and ``how'' co-occurs several times in the training data,
the ISI alignment gives it a high translation probability and aligns it. Our proposed
method also looks at the syntax information of ``thin(g)'' and ``how,'' which are
``Sstatement'' and ``WHNP (WRB),'' respectively. Because this syntax
pair rarely co-occurs in the training data, the learnt weight is negative, which
prevents this incorrect alignment. However, the proposed method incorrectly aligns 
the two ``i'' tokens in AMR and English. This happens because of two reasons:
Firstly, the correct alignment ``:ARG1 i'' has been removed in the preprocessing; 
Secondly, Nile has a feature that tends to align same tokens. 


\begin{table}
\begin{center}
\begin{tabular}{@{}l@{}|r|r@{}} \hline
{\bf Method} & {\bf Our split} & {\bf Original split} \\ \hline \hline
\cite{pust-EtAl:2015:EMNLP} (ISI) & 64.7\% & 65.4\% \\ 
\cite{pust-EtAl:2015:EMNLP} (Proposed) & {\bf 65.1\%} & N/A \\ 
\cite{pust-EtAl:2015:EMNLP} (ISI) w/ rules & N/A & {\bf 67.1\%} \\ \hline
\cite{flanigan-EtAl:2014:P14-1} & 59.0\% & 58.2\% \\
\cite{wang-xue-pradhan:2015:ACL-IJCNLP} & 61.3\% & 63.0\% \\ 
\cite{zhou-EtAl:2016:EMNLP20163} & N/A & 66.0\% \\ \hline
\end{tabular}
\end{center}
\vspace{-3mm}
\caption{\label{table:parsing} Smatch F-scores for AMR parsing.}
\vspace{-2mm}
\end{table}

\subsection{AMR Parsing Results}
The AMR parsing results are shown in Table \ref{table:parsing}, where ''Our split'' 
denotes the performance for different parsers on our data split of the AMR 
corpus.\footnote{As the parsers of \cite{werling-angeli-manning:2015:ACL-IJCNLP,zhou-EtAl:2016:EMNLP20163,Peng:2017:EACL} 
are not publicly available, we could not report the performance of their parser on our data split. We were
not able to run the parsers of \cite{artzi-lee-zettlemoyer:2015:EMNLP,misra-artzi:2016:EMNLP2016} on our split.}
For reference, we also list the parsing accuracies on the original split 
reported in \cite{pust-EtAl:2015:EMNLP,flanigan-EtAl:2014:P14-1,wang-xue-pradhan:2015:ACL-IJCNLP,zhou-EtAl:2016:EMNLP20163}
in the ``Original split'' column of Table \ref{table:parsing}.\footnote{
Note that \cite{flanigan-EtAl:2014:P14-1} did not report the result on this dataset in their paper.
\cite{pust-EtAl:2015:EMNLP} reported the parsing performance of the parser of \cite{flanigan-EtAl:2014:P14-1}
on this dataset, which we listed here.
As \cite{werling-angeli-manning:2015:ACL-IJCNLP,artzi-lee-zettlemoyer:2015:EMNLP,misra-artzi:2016:EMNLP2016}
only reported the parsing performance (62.2\%, 66.2\%, and 66.0\% Smatch F-score, respectively)
on the newswire section of the original split, \cite{Peng:2017:EACL} reported their performance (52\% Smatch F-score)
on a slighly larger dataset (LDC2015E86), we do not list their scores in Table \ref{table:parsing}.}
``\cite{pust-EtAl:2015:EMNLP} (ISI)'' and ``\cite{pust-EtAl:2015:EMNLP} (Proposed)'' denote the parsers using different alignment models.
``\cite{pust-EtAl:2015:EMNLP} (ISI) w/ rules'' denotes a system that further
used rule-based alignments,'' which is not a comparison object in our study.
Note that we cannot report the performance of ``\cite{pust-EtAl:2015:EMNLP} (Proposed)''
on the original split, because our proposed alignment method
requires the 200 pairs in the development and testing sets of the original split for
training and tuning the alignment model.

Parsing accuracies were evaluated using the Smatch F-score \cite{cai-knight:2013:Short}.
As reported \cite{cai-knight:2013:Short}, the Smatch F-score of human inter-annotator is 
in the 79-83 range. We can that see on both our and the original split, \cite{pust-EtAl:2015:EMNLP}
outperforms the other studies.
On our split, the improved alignments by our proposed method
also lead to a 0.4\% Smatch F-score improvement for AMR parsing with the parser of 
\cite{pust-EtAl:2015:EMNLP}. The performance of ``\cite{pust-EtAl:2015:EMNLP} (ISI)'' 
differs on ``Our split'' and ``Original split.'' The reason for this is twofold:
Firstly, ``Original split'' uses 100 more pairs for tuning the parser; 
Secondly, the 100 testing pairs
moved from the testing data of the original split might be more difficult for parsing,
which is consistent with the parsing performance of \cite{wang-xue-pradhan:2015:ACL-IJCNLP}.

\section{Conclusion}
The alignment between English sentences and AMR graphs is necessary for AMR parsing.
We improved the alignment accuracy with a supervised syntax-based alignment method. 
We showed the effectiveness of the supervised method
on both alignment and AMR parsing, even when only a very small training data 
set is available (i.e., 100 pairs).

As future work, firstly, we plan to increase the number of AMR/English sentence 
pairs with gold alignments for training a more accurate alignment model. 
Secondly, we plan to improve the alignment accuracy for roles.
Semantic role labeling \cite{Gildea-EtAl:2000} for the English sentences 
and using the obtained roles for alignment may be a solution for this. 

\section*{Acknowledgments}
We are very appreciated to Mr. Michael Pust for providing the AMR to constituency tree conversion code
and helping conduct the AMR parsing experiments on his parser. We also thank Mr. Yevgeniy Puzikov
very much for helping conduct the AMR parsing experiments on the JAMR and CAMR parsers.


\bibliographystyle{named}
\bibliography{ijcai17}

\end{document}